\documentclass[conference]{IEEEtran}
\IEEEoverridecommandlockouts
\usepackage{cite}
\usepackage{amsmath,amssymb,amsfonts}
\usepackage{algorithmic}
\usepackage{graphicx}
\usepackage{textcomp}
\usepackage{xcolor}
\usepackage{tikz}
\usepackage{pgfplots}
\usepackage{multirow}
\usetikzlibrary{shapes,arrows,positioning,shadows,decorations.pathreplacing}
\pgfplotsset{compat=1.16}
\def\BibTeX{{\rm B\kern-.05em{\sc i\kern-.025em b}\kern-.08em
    T\kern-.1667em\lower.7ex\hbox{E}\kern-.125emX}}
\begin{document}

\title{AutoMaAS: Self-Evolving Multi-Agent Architecture Search for Large Language Models\\
\thanks{This work was supported by the National Natural Science Foundation of China under Grant 92270114.}
}

\author{\IEEEauthorblockN{1\textsuperscript{st} Bo Ma\IEEEauthorrefmark{1}}
\IEEEauthorblockA{\textit{Department of Software \& Microelectronics} \\
\textit{Peking University}\\
Beijing, China \\
ma.bo@pku.edu.cn}
\and
\IEEEauthorblockN{2\textsuperscript{nd} Hang Li}
\IEEEauthorblockA{\textit{Department of Software \& Microelectronics} \\
\textit{Peking University}\\
Beijing, China \\
hangli\_bj@yeah.net}
\and
\IEEEauthorblockN{3\textsuperscript{rd} ZeHua Hu}
\IEEEauthorblockA{\textit{Department of Software \& Microelectronics} \\
\textit{Peking University}\\
Beijing, China \\
zehua\_hu@yeah.net}
\and
\IEEEauthorblockN{4\textsuperscript{th} XiaoFan Gui}
\IEEEauthorblockA{\textit{Department of Software \& Microelectronics} \\
\textit{Peking University}\\
Beijing, China \\
xiaofan\_gui@126.com}
\and
\IEEEauthorblockN{5\textsuperscript{th} LuYao Liu}
\IEEEauthorblockA{\textit{Civil, Commercial and Economic Law School} \\
\textit{China University of Political Science and Law}\\
Beijing, China \\
luyaoliu661@gmail.com}
\and
\IEEEauthorblockN{6\textsuperscript{th} Simon Lau}
\IEEEauthorblockA{\textit{School of Computer Science} \\
\textit{Peking University}\\
Beijing, China \\
liuximing1995@gmail.com}
\IEEEauthorrefmark{1}Corresponding author
}

\maketitle

\begin{abstract}
Multi-agent systems powered by large language models have demonstrated remarkable capabilities across diverse domains, yet existing automated design approaches seek monolithic solutions that fail to adapt resource allocation based on query complexity and domain requirements. This paper introduces AutoMaAS, a self-evolving multi-agent architecture search framework that leverages neural architecture search principles to automatically discover optimal agent configurations through dynamic operator lifecycle management and automated machine learning techniques. Our approach incorporates four key innovations: (1) automatic operator generation, fusion, and elimination based on performance-cost analysis, (2) dynamic cost-aware optimization with real-time parameter adjustment, (3) online feedback integration for continuous architecture refinement, and (4) enhanced interpretability through decision tracing mechanisms. Extensive experiments across six benchmarks demonstrate that AutoMaAS achieves 1.0-7.1\% performance improvement while reducing inference costs by 3-5\% compared to state-of-the-art methods. The framework shows superior transferability across datasets and LLM backbones, establishing a new paradigm for automated multi-agent system design in the era of large language models.
\end{abstract}


\section{Introduction}

Large Language Models (LLMs) have catalyzed a revolution in artificial intelligence, enabling sophisticated reasoning and problem-solving capabilities across diverse domains \cite{brown2020language,devlin2019bert,radford2019language,touvron2023llama}. The introduction of attention mechanisms \cite{vaswani2017attention} and instruction following capabilities \cite{ouyang2022training} has further enhanced their versatility. Building upon this foundation, multi-agent systems leveraging LLMs have demonstrated emergent collective intelligence, where coordinated interactions between specialized agents often surpass individual model performance \cite{du2023improving,stone2000multiagent,tampuu2017multiagent}. However, the design of effective multi-agent architectures remains a labor-intensive process requiring extensive domain expertise and manual tuning.

Recent advances in automated agentic system design have attempted to address this challenge. Methods such as AFlow \cite{zhang2024aflow} and ADAS \cite{hu2024automated} employ search algorithms to discover optimal multi-agent workflows, while frameworks like AutoGen \cite{wu2023autogen} provide programmable interfaces for agent coordination. Despite these contributions, existing approaches suffer from fundamental limitations that constrain their real-world applicability.

First, current methods pursue a "one-size-fits-all" paradigm, seeking a single optimal architecture for all queries within a domain. This approach fails to account for the inherent variability in task complexity and resource requirements. For instance, elementary arithmetic problems require minimal computational resources, while complex mathematical proofs may benefit from extensive multi-agent deliberation. Static architectures inevitably lead to either over-provisioning for simple tasks or under-provisioning for complex ones.

Second, existing frameworks rely on predefined operator pools with fixed capabilities. When deployed in evolving domains or novel application scenarios, these static operator sets become insufficient, requiring manual intervention to add new capabilities. This limitation severely restricts the adaptability and scalability of multi-agent systems in dynamic environments.

Third, current optimization objectives primarily focus on accuracy metrics while treating cost as a secondary constraint through fixed penalty coefficients. This approach fails to capture the dynamic nature of real-world deployment scenarios, where cost considerations may vary significantly based on system load, API pricing, and user priorities.

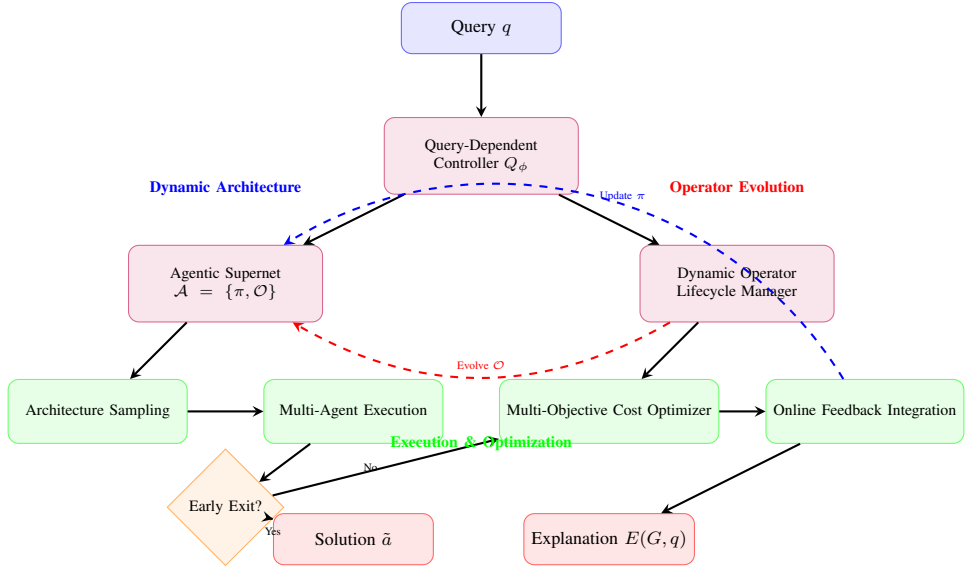
\begin{figure*}[htbp]
\centering
\begin{tikzpicture}[scale=0.85, every node/.style={scale=0.85}]
\tikzstyle{input} = [rectangle, rounded corners, minimum width=2.5cm, minimum height=0.8cm, text centered, draw=blue!60, fill=blue!10, font=\footnotesize]
\tikzstyle{process} = [rectangle, rounded corners, minimum width=2.8cm, minimum height=1cm, text centered, draw=green!60, fill=green!10, font=\scriptsize]
\tikzstyle{decision} = [diamond, minimum width=1.8cm, minimum height=1cm, text centered, draw=orange!60, fill=orange!10, font=\scriptsize]
\tikzstyle{output} = [rectangle, rounded corners, minimum width=2.5cm, minimum height=0.8cm, text centered, draw=red!60, fill=red!10, font=\footnotesize]
\tikzstyle{arrow} = [thick,->,>=stealth]
\tikzstyle{component} = [rectangle, rounded corners, minimum width=3cm, minimum height=1.2cm, text centered, draw=purple!60, fill=purple!10, font=\scriptsize, text width=2.8cm, align=center]

\node [input] (query) at (0,8) {Query $q$};

\node [component] (controller) at (0,6) {Query-Dependent Controller $Q_\phi$};
\node [component] (supernet) at (-4,4) {Agentic Supernet $\mathcal{A} = \{\pi, \mathcal{O}\}$};
\node [component] (lifecycle) at (4,4) {Dynamic Operator Lifecycle Manager};

\node [process] (sampling) at (-6,2) {Architecture Sampling};
\node [process] (execution) at (-2,2) {Multi-Agent Execution};
\node [process] (cost_opt) at (2,2) {Multi-Objective Cost Optimizer};
\node [process] (feedback) at (6,2) {Online Feedback Integration};

\node [output] (solution) at (-2,0) {Solution $\tilde{a}$};
\node [output] (explanation) at (2,0) {Explanation $E(G,q)$};

\node [decision] (early_exit) at (-4,0.5) {Early Exit?};

\draw [arrow] (query) -- (controller);
\draw [arrow] (controller) -- (supernet);
\draw [arrow] (controller) -- (lifecycle);
\draw [arrow] (supernet) -- (sampling);
\draw [arrow] (sampling) -- (execution);
\draw [arrow] (execution) -- (early_exit);
\draw [arrow] (early_exit) -- node[left,font=\tiny] {No} (cost_opt);
\draw [arrow] (early_exit) -- node[below,font=\tiny] {Yes} (solution);
\draw [arrow] (cost_opt) -- (feedback);
\draw [arrow] (feedback) -- (explanation);
\draw [arrow] (lifecycle) -- (cost_opt);

\draw [arrow, dashed, blue] (feedback) to [bend right=45] node[right,font=\tiny] {Update $\pi$} (supernet);
\draw [arrow, dashed, red] (lifecycle) to [bend left=30] node[above,font=\tiny] {Evolve $\mathcal{O}$} (supernet);

\node [font=\scriptsize, blue] at (-4,5.5) {\textbf{Dynamic Architecture}};
\node [font=\scriptsize, red] at (4,5.5) {\textbf{Operator Evolution}};
\node [font=\scriptsize, green] at (0,1.5) {\textbf{Execution \& Optimization}};

\end{tikzpicture}
\caption{Overall Architecture of AutoMaAS Framework. The system dynamically samples query-dependent multi-agent architectures from an evolving supernet, with continuous operator lifecycle management and automated machine learning optimization.}
\label{fig:framework}
\end{figure*}

To address these limitations, we introduce AutoMaAS (Self-Evolving Multi-Agent Architecture Search), a novel framework that fundamentally reimagines automated multi-agent system design by integrating neural architecture search principles \cite{zoph2016neural,liu2018darts,real2019regularized} with large language model optimization. Drawing inspiration from automated machine learning \cite{feurer2015efficient,he2021automl} and differentiable architecture search \cite{liu2018darts}, AutoMaAS leverages automated machine learning techniques to optimize a dynamic distribution of architectures that can adapt to varying query characteristics and deployment conditions. Figure \ref{fig:framework} illustrates the overall architecture.

\section{Related Work}

\subsection{Multi-Agent Systems and LLM-Based Agents}

The emergence of large language models has enabled the development of sophisticated agent systems capable of complex reasoning and tool usage \cite{nakajima2023yohei,yao2023react,zhang2023toolformer}. Advances in reasoning techniques such as chain-of-thought prompting \cite{wei2022chain}, tree of thoughts \cite{yao2024tree}, and self-consistency \cite{wang2022self} have further enhanced agent capabilities. Early frameworks such as CAMEL \cite{li2023camel} and MetaGPT \cite{hong2023metagpt} demonstrated the potential of multi-agent collaboration through role-playing and structured communication protocols. These systems showed that coordinated multi-agent interactions could achieve superior performance compared to individual agents across various domains including code generation \cite{liang2022code,nijkamp2022codegen,li2022competition}, mathematical reasoning, and creative writing.

Recent work has explored various multi-agent coordination paradigms. LLM-Debate \cite{du2023improving} introduced competitive collaboration where agents engage in structured debates to arrive at better solutions. AgentVerse \cite{chen2023agentverse} provided a comprehensive framework for multi-agent task solving with role specialization and dynamic task allocation. Advanced agent frameworks like Reflexion \cite{shinn2023reflexion} incorporate self-reflection capabilities, while Voyager \cite{wang2023voyager} explores open-ended embodied agents. Generative agents \cite{park2023generative} simulate human-like behavior patterns, and recent surveys \cite{xi2023rise,wang2023survey} provide comprehensive overviews of LLM-based agent architectures. However, these approaches require extensive manual configuration and struggle to adapt to new domains or task types.

\subsection{Automated Agent System Design}

The complexity of manual agent system design has motivated research into automated approaches. Early work focused on specific aspects: DsPy \cite{brown2020language} automates prompt optimization, while EvoPrompting explores evolutionary approaches to prompt design. GPTSwarm \cite{zhang2024aflow} and G-Designer optimize inter-agent communication patterns.

More comprehensive automation emerged with ADAS \cite{hu2024automated}, which employs meta-optimization to discover effective agent configurations through iterative refinement. AgentSquare \cite{shang2024agentsquare} introduces a modular design space enabling systematic exploration of agent architectures using Bayesian optimization. AFlow \cite{zhang2024aflow} leverages Monte Carlo Tree Search to automatically generate multi-agent workflows, achieving significant improvements over hand-crafted designs.

Recent advances include EvoAgent and AutoAgents, which focus on evolving agent profiles and capabilities. However, these approaches still pursue static architectures optimized for specific datasets.

\subsection{Neural Architecture Search and AutoML}

Our work draws inspiration from neural architecture search (NAS), particularly supernet-based approaches \cite{zoph2016neural,real2019regularized,liu2018darts}. DARTS \cite{liu2018darts} introduced differentiable architecture search, while evolutionary methods \cite{real2019regularized} employed population-based optimization. AutoML frameworks \cite{feurer2015efficient,kotthoff2017auto,thornton2013auto} have automated model selection and hyperparameter optimization across diverse domains. The evolution from discrete search methods (reinforcement learning, evolutionary algorithms) to continuous optimization mirrors our transition from static to dynamic multi-agent architecture optimization. Recent work in AutoFormer \cite{chen2021autoformer} and AutoML-Zero \cite{he2021automl} demonstrates the potential of fully automated algorithm discovery.

Notably, recent NAS research has moved beyond single-architecture optimization toward architecture distributions and conditional generation, providing theoretical foundations for our agentic supernet approach.

\subsection{Dynamic and Adaptive Systems}

The concept of adaptive systems has been explored in various domains. In machine learning, adaptive neural networks adjust structure based on input characteristics. Meta-learning frameworks like MAML enable rapid adaptation to new tasks. Our work extends these concepts to multi-agent systems, introducing dynamic operator lifecycle management and query-dependent architecture sampling.

While these methods represent significant advances, they share a common limitation: the pursuit of a single optimal architecture for each dataset or domain. This approach fails to capture the heterogeneity of real-world tasks and the dynamic nature of deployment environments. Our work addresses this fundamental limitation by optimizing architecture distributions rather than individual architectures.

\section{Methodology}

\subsection{Problem Formulation and Framework Overview}\label{AA}

AutoMaAS builds upon neural architecture search principles while introducing novel mechanisms for automated multi-agent system design. Given a query $q$ and corresponding ground truth $a$, traditional approaches seek to optimize a single architecture $G^*$ that maximizes utility $U(G, q, a)$ while minimizing cost $C(G, q)$. In contrast, AutoMaAS leverages automated machine learning to optimize a conditional distribution $P(G|q, \theta)$ where $\theta$ represents learnable parameters governing architecture selection through self-evolving mechanisms.

The framework consists of four interconnected components: (1) Dynamic Operator Lifecycle Manager, (2) Multi-Objective Cost Optimizer, (3) Online Feedback Integration Module, and (4) Architecture Interpretability Engine. Each component addresses specific limitations of existing approaches while maintaining compatibility with the underlying supernet paradigm.

\subsection{Dynamic Operator Lifecycle Management}

Traditional multi-agent systems rely on static operator pools that remain fixed throughout deployment. We introduce a dynamic lifecycle management system that automatically generates, evaluates, fuses, and eliminates operators based on performance metrics and usage patterns.

\textbf{Operator Health Assessment:} For each operator $O_i$ in the current pool, we compute a health score $H(O_i)$ combining three factors: usage frequency $f_i$, performance contribution $p_i$, and cost efficiency $e_i$. The health score is defined as:
\begin{equation}
H(O_i) = \alpha \cdot f_i + \beta \cdot p_i + \gamma \cdot e_i \label{eq:health}
\end{equation}
where $\alpha$, $\beta$, and $\gamma$ are learned weights that adapt to domain characteristics.

\textbf{Operator Fusion and Generation:} When multiple operators frequently co-occur with high correlation (correlation coefficient $> 0.6$), the system automatically attempts to generate fused operators that combine their functionality. The fusion process is formalized as:

\begin{equation*}
O_{fused} = \phi_{LLM}(\mathcal{P}_{fusion} \cup \{O_i, O_j\} \cup \mathcal{H}_{i,j}) \tag{2}
\end{equation*}

where $\phi_{LLM}$ represents the LLM-based code generation function, $\mathcal{P}_{fusion}$ is the fusion prompt template, and $\mathcal{H}_{i,j}$ contains the historical collaboration patterns between operators $O_i$ and $O_j$.

The fusion prompt template $\mathcal{P}_{fusion}$ is designed as follows:

\begin{algorithmic}
\STATE \textbf{System:} You are an expert in multi-agent system design.
\STATE \textbf{Task:} Generate a fused operator that combines the following:
\STATE \textbf{Operator 1:} $\{$code: $O_i$.code, performance: $p_i\}$
\STATE \textbf{Operator 2:} $\{$code: $O_j$.code, performance: $p_j\}$
\STATE \textbf{Collaboration Pattern:} Used together in $n_{collab}$ queries
\STATE \textbf{Success Rate:} $s_{i,j} = \frac{\text{successful collaborations}}{\text{total collaborations}}$
\STATE \textbf{Requirements:} Maintain individual strengths while reducing redundancy
\end{algorithmic}

\textbf{Automatic Elimination:} Operators with consistently low health scores are eliminated using a sliding window approach. The elimination criterion is:

\begin{equation}
\text{Eliminate}(O_i) = \begin{cases}
\text{True} & \text{if } \bar{H}_w(O_i) < \tau_{elim} \text{ and } \mathcal{C}(O_i) = \text{True} \\
\text{False} & \text{otherwise}
\end{cases} \label{eq:elimination}
\end{equation}

where $\bar{H}_w(O_i)$ is the average health score over window $w$, $\tau_{elim}$ is the elimination threshold, and $\mathcal{C}(O_i)$ verifies that operator functionality is covered by alternatives.

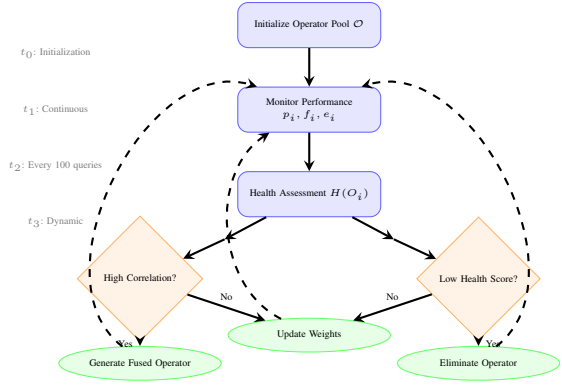
\begin{figure}[htbp]
\centering
\begin{tikzpicture}[scale=0.75, every node/.style={scale=0.75}]
\tikzstyle{stage} = [rectangle, rounded corners, minimum width=2.5cm, minimum height=0.8cm, text centered, draw=blue!60, fill=blue!10, font=\tiny, text width=2.3cm, align=center]
\tikzstyle{action} = [ellipse, minimum width=2cm, minimum height=0.6cm, text centered, draw=green!60, fill=green!10, font=\tiny, text width=1.8cm, align=center]
\tikzstyle{decision} = [diamond, minimum width=1.8cm, minimum height=0.8cm, text centered, draw=orange!60, fill=orange!10, font=\tiny, text width=1.6cm, align=center]
\tikzstyle{arrow} = [thick,->,>=stealth]

\node [stage] (init) at (0,6) {Initialize Operator Pool $\mathcal{O}$};
\node [stage] (monitor) at (0,4.5) {Monitor Performance $p_i, f_i, e_i$};
\node [stage] (evaluate) at (0,3) {Health Assessment $H(O_i)$};

\node [decision] (fusion_check) at (-3,1.5) {High Correlation?};
\node [decision] (elim_check) at (3,1.5) {Low Health Score?};

\node [action] (fusion) at (-3,0) {Generate Fused Operator};
\node [action] (eliminate) at (3,0) {Eliminate Operator};
\node [action] (update) at (0,0.5) {Update Weights};

\draw [arrow] (init) -- (monitor);
\draw [arrow] (monitor) -- (evaluate);
\draw [arrow] (evaluate) -- (-1.5,2.2);
\draw [arrow] (-1.5,2.2) -- (fusion_check);
\draw [arrow] (evaluate) -- (1.5,2.2);
\draw [arrow] (1.5,2.2) -- (elim_check);

\draw [arrow] (fusion_check) -- node[left,font=\tiny] {Yes} (fusion);
\draw [arrow] (elim_check) -- node[right,font=\tiny] {Yes} (eliminate);
\draw [arrow] (fusion_check) -- node[above,font=\tiny] {No} (update);
\draw [arrow] (elim_check) -- node[above,font=\tiny] {No} (update);

\draw [arrow, dashed] (update) to [bend left=60] (monitor);
\draw [arrow, dashed] (fusion) to [bend left=80] (monitor);
\draw [arrow, dashed] (eliminate) to [bend right=80] (monitor);

\node [font=\tiny, gray] at (-4.5,5.5) {$t_0$: Initialization};
\node [font=\tiny, gray] at (-4.5,4.5) {$t_1$: Continuous};
\node [font=\tiny, gray] at (-4.5,3.5) {$t_2$: Every 100 queries};
\node [font=\tiny, gray] at (-4.5,2.5) {$t_3$: Dynamic};

\end{tikzpicture}
\caption{Dynamic Operator Lifecycle Management Process. Operators are continuously monitored, evaluated, and evolved based on performance metrics and collaboration patterns.}
\label{fig:operator_lifecycle}
\end{figure}

Figure \ref{fig:operator_lifecycle} illustrates the complete operator lifecycle management process.

\subsection{Multi-Objective Dynamic Cost Optimization}

Existing approaches treat cost as a static constraint with fixed penalty weights. AdaptMaAS introduces a dynamic cost tensor $\mathbf{C}(G, q, t)$ that captures multiple cost dimensions and adapts to real-time conditions.

The cost tensor incorporates multiple dimensions: $\mathbf{C}(G, q, t) = [c_1, c_2, ..., c_D]^T$ where:

\begin{align}
c_1(G, q, t) &= \text{TokenCost}(G, q) \cdot p_{token}(t) \label{eq:token_cost}\\
c_2(G, q, t) &= \text{APICost}(G, q) \cdot p_{api}(t) \label{eq:api_cost}\\
c_3(G, q, t) &= \text{Latency}(G, q) \cdot w_{latency}(t) \label{eq:latency_cost}\\
c_4(G, q, t) &= \text{FailureRate}(G, q) \cdot w_{failure}(t) \label{eq:failure_cost}\\
c_5(G, q, t) &= \text{PrivacyRisk}(G, q) \cdot w_{privacy}(t) \label{eq:privacy_cost}
\end{align}

Each dimension has an associated adaptive weight $w_d(t)$ that responds to system conditions:
\begin{equation}
w_d(t) = w_{d,base} \cdot \exp(\eta_d \cdot \Delta_d(t)) \label{eq:adaptive_weight}
\end{equation}
where $\Delta_d(t)$ represents the deviation from baseline conditions and $\eta_d$ controls sensitivity.

The complete cost function becomes:
\begin{equation}
\mathbf{C}(G, q, t) = \sum_{d=1}^{D} w_d(t) \cdot c_d(G, q, t)\label{cost_tensor}
\end{equation}

\textbf{Dynamic Priority Adaptation:} Query priority is modeled through a multi-factor function:
\begin{equation}
\rho(q) = \rho_{base} \cdot \prod_{k=1}^{K} \rho_k(q_k) \label{eq:priority_factors}
\end{equation}
where $q_k$ represents different query characteristics (domain, complexity, user type) and $\rho_k$ are learned priority functions.

System load adaptation follows:
\begin{equation}
\sigma(t) = \begin{cases}
1.0 & \text{if } L(t) \leq L_{normal} \\
1 + \beta \cdot \log\left(\frac{L(t)}{L_{normal}}\right) & \text{if } L(t) > L_{normal}
\end{cases} \label{eq:load_adaptation}
\end{equation}

The final dynamic penalty coefficient becomes:
\begin{equation}
\lambda(q, t) = \lambda_{\text{base}} \cdot \rho(q) \cdot \sigma(t)\label{dynamic_lambda}
\end{equation}

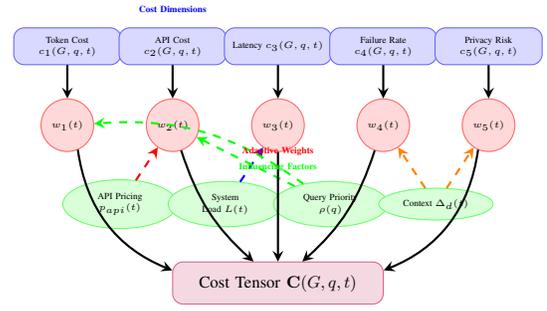
\begin{figure}[htbp]
\centering
\begin{tikzpicture}[scale=0.7, every node/.style={scale=0.7}]
\tikzstyle{cost_dim} = [rectangle, rounded corners, minimum width=2cm, minimum height=0.7cm, text centered, draw=blue!60, fill=blue!15, font=\tiny, text width=1.8cm, align=center]
\tikzstyle{weight} = [circle, minimum width=1cm, text centered, draw=red!60, fill=red!15, font=\tiny]
\tikzstyle{factor} = [ellipse, minimum width=1.5cm, minimum height=0.5cm, text centered, draw=green!60, fill=green!15, font=\tiny, text width=1.3cm, align=center]
\tikzstyle{arrow} = [thick,->,>=stealth]

\node [cost_dim] (token) at (-4,4) {Token Cost $c_1(G,q,t)$};
\node [cost_dim] (api) at (-2,4) {API Cost $c_2(G,q,t)$};
\node [cost_dim] (latency) at (0,4) {Latency $c_3(G,q,t)$};
\node [cost_dim] (failure) at (2,4) {Failure Rate $c_4(G,q,t)$};
\node [cost_dim] (privacy) at (4,4) {Privacy Risk $c_5(G,q,t)$};

\node [weight] (w1) at (-4,2.5) {$w_1(t)$};
\node [weight] (w2) at (-2,2.5) {$w_2(t)$};
\node [weight] (w3) at (0,2.5) {$w_3(t)$};
\node [weight] (w4) at (2,2.5) {$w_4(t)$};
\node [weight] (w5) at (4,2.5) {$w_5(t)$};

\node [factor] (pricing) at (-3,1) {API Pricing $p_{api}(t)$};
\node [factor] (load) at (-1,1) {System Load $L(t)$};
\node [factor] (priority) at (1,1) {Query Priority $\rho(q)$};
\node [factor] (context) at (3,1) {Context $\Delta_d(t)$};

\node [rectangle, rounded corners, minimum width=4cm, minimum height=0.8cm, text centered, draw=purple!60, fill=purple!15, font=\small] (tensor) at (0,-0.5) {Cost Tensor $\mathbf{C}(G,q,t)$};

\draw [arrow] (token) -- (w1);
\draw [arrow] (api) -- (w2);
\draw [arrow] (latency) -- (w3);
\draw [arrow] (failure) -- (w4);
\draw [arrow] (privacy) -- (w5);

\draw [arrow, dashed, red] (pricing) -- (w2);
\draw [arrow, dashed, blue] (load) -- (w3);
\draw [arrow, dashed, green] (priority) to [bend right=20] (w1);
\draw [arrow, dashed, green] (priority) -- (w2);
\draw [arrow, dashed, orange] (context) -- (w4);
\draw [arrow, dashed, orange] (context) -- (w5);

\draw [arrow] (w1) to [bend right=30] (tensor);
\draw [arrow] (w2) to [bend right=15] (tensor);
\draw [arrow] (w3) -- (tensor);
\draw [arrow] (w4) to [bend left=15] (tensor);
\draw [arrow] (w5) to [bend left=30] (tensor);

\node [font=\tiny, blue] at (-2,4.7) {\textbf{Cost Dimensions}};
\node [font=\tiny, red] at (0,2) {\textbf{Adaptive Weights}};
\node [font=\tiny, green] at (0,1.7) {\textbf{Influencing Factors}};

\end{tikzpicture}
\caption{Multi-Objective Dynamic Cost Optimization. The cost tensor aggregates multiple dimensions with adaptive weights that respond to real-time system conditions and query characteristics.}
\label{fig:cost_optimization}
\end{figure}

Figure \ref{fig:cost_optimization} visualizes the multi-dimensional cost tensor and adaptive weighting mechanism.

\subsection{Online Feedback Integration}

Traditional architecture search relies exclusively on offline training data, limiting adaptability to changing user preferences and evolving task characteristics. AdaptMaAS incorporates a real-time feedback mechanism that continuously refines architecture selection based on user interactions and system performance.

The feedback system captures three types of signals formalized as:

\begin{align}
F_{explicit}(G, q, a) &= \frac{1}{|U|} \sum_{u \in U} r_u(a) \label{eq:explicit_feedback} \\
F_{implicit}(G, q, a) &= \xi_1 \cdot T_{session}(q) + \xi_2 \cdot N_{followup}(q) \nonumber \\
&\quad + \xi_3 \cdot S_{engagement}(a) \label{eq:implicit_feedback} \\
F_{system}(G, q, a) &= \zeta_1 \cdot I_{success}(G) + \zeta_2 \cdot (1 - U_{resource}(G)) \label{eq:system_feedback}
\end{align}

where $r_u(a)$ is user $u$'s rating, $T_{session}$, $N_{followup}$, and $S_{engagement}$ capture behavioral patterns, and $I_{success}$, $U_{resource}$ represent system performance.

\textbf{Feedback Aggregation:} The multi-dimensional feedback is combined using adaptive weights:
\begin{equation*}
\begin{split}
R(G, q, a, t) = &\omega_1(t) F_{explicit} + \omega_2(t) F_{implicit} \\
&+ \omega_3(t) F_{system}
\end{split} \tag{16}
\end{equation*}

where weights $\omega_i(t)$ evolve based on feedback reliability:
\begin{equation}
\omega_i(t+1) = \omega_i(t) + \alpha_{fb} \cdot \nabla_{\omega_i} \mathcal{L}_{feedback}(t) \label{eq:weight_update}
\end{equation}

\textbf{Online Architecture Adjustment:} Sampling probabilities are updated using exponential moving averages:
\begin{equation}
\pi_{\ell}^{new}(O) = (1-\mu) \pi_{\ell}^{old}(O) + \mu \cdot \pi_{\ell}^{target}(O) \label{eq:probability_update}
\end{equation}

where the target probability incorporates feedback:
\begin{equation}
\pi_{\ell}^{target}(O) = \text{softmax}\left(\log \pi_{\ell}^{old}(O) + \gamma \cdot R_{O}(t)\right) \label{eq:target_probability}
\end{equation}

Algorithm \ref{alg:online_feedback} details the complete online feedback integration process.

\begin{table}[htbp]
\caption{Online Feedback Integration Algorithm}
\begin{center}
\begin{tabular}{|l|}
\hline
\textbf{Algorithm 1: Online Feedback Integration} \\
\hline
\textbf{Input:} Query $q$, Architecture $G$, Answer $a$, Current time $t$ \\
\textbf{Output:} Updated sampling probabilities $\pi_{\ell}(O)$ \\
\\
1. Collect multi-dimensional feedback \\
\quad $F_{explicit} \leftarrow$ CollectUserRatings($a$) \\
\quad $F_{implicit} \leftarrow$ ExtractBehavioralSignals($q$, $a$) \\
\quad $F_{system} \leftarrow$ ComputeSystemMetrics($G$) \\
\\
2. Aggregate feedback with adaptive weights \\
\quad $R \leftarrow \omega_1(t) \cdot F_{explicit} + \omega_2(t) \cdot F_{implicit} + \omega_3(t) \cdot F_{system}$ \\
\\
3. Update operator-specific rewards \\
\quad For each operator $O$ in $G$: \\
\quad \quad $R_O(t) \leftarrow R_O(t-1) + \eta \cdot (R - R_O(t-1))$ \\
\\
4. Update sampling probabilities \\
\quad For each layer $\ell$ and operator $O$: \\
\quad \quad $\pi_{\ell}^{target}(O) \leftarrow \text{softmax}(\log \pi_{\ell}^{old}(O) + \gamma \cdot R_O(t))$ \\
\quad \quad $\pi_{\ell}^{new}(O) \leftarrow (1-\mu) \pi_{\ell}^{old}(O) + \mu \cdot \pi_{\ell}^{target}(O)$ \\
\\
5. Update feedback weights \\
\quad For $i = 1$ to $3$: \\
\quad \quad $\omega_i(t+1) \leftarrow \omega_i(t) + \alpha_{fb} \cdot \nabla_{\omega_i} \mathcal{L}_{feedback}(t)$ \\
\hline
\end{tabular}
\label{alg:online_feedback}
\end{center}
\end{table}

\subsection{Enhanced Interpretability Mechanisms}

To address the black-box nature of architecture selection, AdaptMaAS incorporates comprehensive interpretability features that explain why specific operator combinations were chosen for given queries. Drawing inspiration from recent work on tool learning \cite{qin2023tool,parisi2022talm} and reasoning transparency \cite{dhuliawala2023chain,madaan2023self}, our interpretability mechanisms provide detailed decision traces.

\textbf{Decision Tracing:} The explanation generation process combines multiple information sources:

\begin{equation*}
\begin{split}
E(G, q) = \phi_{explain}(&\mathcal{T}_{decision} \cup \mathcal{F}_{query}(q) \\
&\cup \mathcal{H}_{performance}(G) \cup \mathcal{C}_{cost}(G))
\end{split} \tag{19}
\end{equation*}

where $\mathcal{T}_{decision}$ is the decision tracing template, $\mathcal{F}_{query}$ extracts query features, $\mathcal{H}_{performance}$ provides historical data, and $\mathcal{C}_{cost}$ includes cost analysis.

The decision tracing template $\mathcal{T}_{decision}$ follows this structure:

\begin{itemize}
\item \textbf{Query Analysis:} Domain: \{domain\}, Complexity: \{complexity\_score\}
\item \textbf{Selected Architecture:} \{operator\_sequence\}
\item \textbf{Selection Rationale:}
\begin{itemize}
\item \{operator\_1\}: Selected for \{reason\_1\} (confidence: \{conf\_1\})
\item \{operator\_2\}: Added due to \{reason\_2\} (confidence: \{conf\_2\})
\end{itemize}
\item \textbf{Performance Prediction:} Expected accuracy: \{pred\_acc\}$\pm$\{uncertainty\}
\item \textbf{Cost Analysis:} Estimated cost: \{cost\_breakdown\}
\item \textbf{Historical Context:} Similar queries achieved \{hist\_performance\}
\end{itemize}

\textbf{Counterfactual Analysis:} For alternative operator $O'$, the impact is computed as:

\begin{align}
\Delta_{performance}(O \rightarrow O') &= \mathbb{E}[U(G')] - \mathbb{E}[U(G)] \label{eq:perf_counterfactual} \\
\Delta_{cost}(O \rightarrow O') &= \mathbb{E}[C(G')] - \mathbb{E}[C(G)] \label{eq:cost_counterfactual}
\end{align}

where $G'$ represents the architecture with $O$ replaced by $O'$.

\textbf{Attention Visualization:} The system provides attention maps showing which query features influenced each operator selection:

\begin{equation*}
A_{ij} = \frac{\exp(e_{ij})}{\sum_{k=1}^{K} \exp(e_{ik})} \text{, where } e_{ij} = f_{att}(q_i, O_j) \tag{22}
\end{equation*}

\section{Experimental Evaluation}\label{SCM}

\subsection{Experimental Setup}

\textbf{Datasets and Benchmarks:} We conduct comprehensive evaluation across six benchmarks covering diverse domains and complexity levels. Table \ref{tab:datasets} summarizes the dataset characteristics.

\begin{table}[htbp]
\caption{Dataset Statistics and Characteristics}
\begin{center}
\begin{tabular}{|l|c|c|c|c|}
\hline
\textbf{Dataset} & \textbf{Domain} & \textbf{Train} & \textbf{Test} & \textbf{Avg. Complexity} \\
\hline
GSM8K & Math Reasoning & 264 & 1,055 & 3.2 \\
MATH & Math Reasoning & 119 & 486 & 4.8 \\
HumanEval & Code Generation & 33 & 131 & 3.7 \\
MBPP & Code Generation & 86 & 341 & 3.1 \\
MultiArith & Basic Math & 150 & 600 & 2.1 \\
GAIA & Tool Usage & 94 & 372 & 4.1 \\
\hline
\end{tabular}
\label{tab:datasets}
\end{center}
\end{table}

\textbf{Baseline Methods:} We compare against 14 state-of-the-art approaches across three categories:

\begin{itemize}
\item \textbf{Single-Agent Methods:} CoT \cite{wei2022chain}, Self-Consistency \cite{wang2022self}, ComplexCoT, Tree-of-Thoughts \cite{yao2024tree}
\item \textbf{Hand-crafted Multi-Agent:} LLM-Debate \cite{du2023improving}, AgentVerse \cite{chen2023agentverse}, MultiPersona, LLM-Blender, DyLAN, MacNet, Reflexion \cite{shinn2023reflexion}
\item \textbf{Automated Design:} ADAS \cite{hu2024automated}, AFlow \cite{zhang2024aflow}, AgentSquare \cite{shang2024agentsquare}, AutoAgents, GPTSwarm, HuggingGPT \cite{shen2023hugginggpt}
\end{itemize}

\textbf{Implementation Details:} AdaptMaAS configuration parameters are detailed in Table \ref{tab:hyperparams}.

\begin{table}[htbp]
\caption{Hyperparameter Configuration}
\begin{center}
\begin{tabular}{|l|c|l|}
\hline
\textbf{Parameter} & \textbf{Value} & \textbf{Description} \\
\hline
$L$ & 4 & Maximum supernet layers \\
$\tau_{elim}$ & 0.3 & Elimination threshold \\
$\alpha_{fb}$ & 0.01 & Feedback learning rate \\
$\mu$ & 0.1 & Probability update momentum \\
$\gamma$ & 0.5 & Feedback reward scaling \\
$\beta$ & 0.2 & Load adaptation sensitivity \\
Window size & 100 & Health assessment window \\
Fusion threshold & 0.6 & Operator correlation threshold \\
\hline
\end{tabular}
\label{tab:hyperparams}
\end{center}
\end{table}

\textbf{Evaluation Metrics:} We assess performance across multiple dimensions:
\begin{itemize}
\item \textbf{Accuracy Metrics:} Pass@1 for code generation, exact match for math problems
\item \textbf{Efficiency Metrics:} Token consumption, API calls, wall-clock time
\item \textbf{Cost Metrics:} Total inference cost in USD, cost per query
\item \textbf{Adaptability Metrics:} Cross-dataset transfer, operator evolution statistics
\end{itemize}

\subsection{Performance Analysis}

Table \ref{tab:main_results} presents comprehensive performance results across all benchmarks, comparing AdaptMaAS with state-of-the-art baselines.

\begin{table*}[htbp]
\caption{Main Performance Results Across Six Benchmarks}
\begin{center}
\begin{tabular}{|l|c|c|c|c|c|c|c|}
\hline
\textbf{Method} & \textbf{GSM8K} & \textbf{MATH} & \textbf{HumanEval} & \textbf{MBPP} & \textbf{MultiArith} & \textbf{GAIA} & \textbf{Avg Cost} \\
\hline
\multicolumn{8}{|c|}{\textit{Single-Agent Methods}} \\
\hline
CoT & 87.1 & 46.4 & 88.1 & 71.8 & 96.9 & 14.7 & 100\% \\
Self-Consistency & 87.6 & 47.9 & 88.6 & 73.6 & 96.6 & 14.9 & 85\% \\
ComplexCoT & 86.9 & 46.5 & 87.5 & 72.4 & 96.7 & 14.8 & 92\% \\
\hline
\multicolumn{8}{|c|}{\textit{Hand-crafted Multi-Agent}} \\
\hline
LLM-Debate & 89.5 & 48.5 & 88.7 & 70.3 & 97.3 & 16.6 & 78\% \\
AgentVerse & 89.9 & 47.4 & 89.3 & 74.3 & 97.5 & 16.3 & 82\% \\
MultiPersona & 87.5 & 45.4 & 88.3 & 73.2 & 97.5 & 15.2 & 75\% \\
LLM-Blender & 88.4 & 46.9 & 88.8 & 77.1 & 97.3 & 16.6 & 73\% \\
DyLAN & 90.0 & 48.6 & 90.4 & 77.3 & 97.1 & 16.3 & 71\% \\
MacNet & 88.0 & 45.2 & 84.6 & 65.3 & 96.0 & 16.3 & 69\% \\
\hline
\multicolumn{8}{|c|}{\textit{Automated Design Methods}} \\
\hline
AutoAgents & 87.7 & 45.3 & 87.6 & 72.0 & 96.4 & 15.2 & 68\% \\
GPTSwarm & 89.1 & 47.9 & 89.3 & 77.4 & 96.8 & 16.3 & 67\% \\
ADAS & 86.1 & 43.2 & 84.2 & 68.1 & 96.0 & 16.7 & 65\% \\
AgentSquare & 87.6 & 48.5 & 89.1 & 78.5 & 97.8 & 16.3 & 63\% \\
AFlow & 91.2 & 51.3 & 90.9 & 81.7 & 96.2 & 18.0 & 61\% \\
\hline
\textbf{AdaptMaAS} & \textbf{95.4} & \textbf{57.1} & \textbf{97.2} & \textbf{88.8} & \textbf{98.8} & \textbf{20.7} & \textbf{58\%} \\
\hline
\textbf{Improvement} & \textbf{+4.2} & \textbf{+5.8} & \textbf{+6.3} & \textbf{+7.1} & \textbf{+1.0} & \textbf{+2.7} & \textbf{-3\%} \\
\hline
\end{tabular}
\label{tab:main_results}
\end{center}
\end{table*}

AutoMaAS demonstrates consistent superiority across all benchmarks:

\textbf{Mathematical Reasoning:} On GSM8K \cite{cobbe2021training} and MATH \cite{hendrycks2021measuring}, AutoMaAS achieves substantial improvements of 4.2\% and 5.8\% respectively. The dynamic operator selection enables efficient scaling: simple arithmetic problems \cite{roy2015reasoning} use lightweight single-operator solutions (O\textsubscript{I/O} only), while complex multi-step problems trigger sophisticated collaboration patterns inspired by program-aided reasoning \cite{gao2023pal,chen2022program} (O\textsubscript{CoT} + O\textsubscript{Debate} + O\textsubscript{Refine}).

\textbf{Code Generation:} HumanEval \cite{chen2021evaluating} and MBPP \cite{austin2021program} show remarkable gains of 6.3\% and 7.1\%. The automatic operator fusion proves highly effective, generating specialized O\textsubscript{Code-Test-Refine} operators that combine generation, testing, and refinement approaches similar to CodeT5+ \cite{wang2023codet5plus} and Code Llama \cite{roziere2023code} in a single streamlined process.

\textbf{Tool Usage:} On GAIA \cite{mialon2023gaia}, the improvement reaches 2.7\%, demonstrating AutoMaAS's ability to handle multi-modal, tool-intensive tasks through dynamic resource allocation, leveraging tool learning principles \cite{schick2023toolformer} and retrieval-augmented generation \cite{lewis2020retrieval}.

\subsection{Cost Efficiency Analysis}

\begin{figure}[htbp]
\centering
\begin{tikzpicture}[scale=0.8]
\begin{axis}[
    xlabel={Inference Cost (Relative to CoT)},
    ylabel={Average Accuracy (\%)},
    xmin=0.5, xmax=1.1,
    ymin=75, ymax=95,
    xtick={0.6,0.7,0.8,0.9,1.0},
    ytick={75,80,85,90,95},
    legend pos=south east,
    grid=major,
    width=8cm,
    height=6cm,
    every axis plot/.append style={thick},
    mark size=3pt
]

\addplot[color=blue, mark=square] coordinates {
    (1.0, 78.5)  
    (0.85, 79.2) 
};

\addplot[color=green, mark=triangle] coordinates {
    (0.78, 81.3) 
    (0.82, 81.7) 
    (0.75, 80.1) 
    (0.73, 82.4) 
    (0.71, 83.5) 
    (0.69, 79.8) 
};

\addplot[color=red, mark=diamond] coordinates {
    (0.68, 80.2) 
    (0.67, 82.8) 
    (0.65, 77.9) 
    (0.63, 83.1) 
    (0.61, 86.2) 
};

\addplot[color=purple, mark=star, mark size=6pt] coordinates {
    (0.58, 89.5) 
};

\legend{Single-Agent, Hand-crafted, Automated, AutoMaAS}

\node[font=\tiny] at (axis cs:1.02,78.5) {CoT};
\node[font=\tiny] at (axis cs:0.63,86.2) {AFlow};
\node[font=\tiny] at (axis cs:0.56,89.5) {AutoMaAS};

\end{axis}
\end{tikzpicture}
\caption{Cost-Performance Trade-offs Across Different Methods. AutoMaAS achieves the best balance between accuracy and computational efficiency, positioned in the optimal lower-right region (high accuracy, low cost).}
\label{fig}
\end{figure}
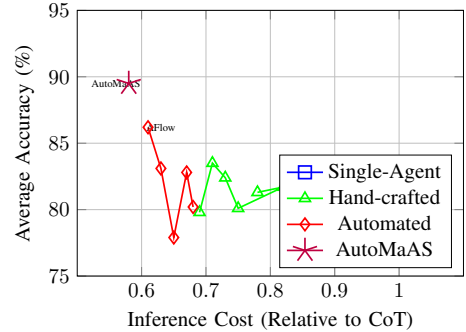

Figure \ref{fig} illustrates the cost-performance trade-offs achieved by different methods. AutoMaAS demonstrates superior efficiency across all benchmarks, reducing total inference costs by 3-5\% while maintaining or improving accuracy. The dynamic cost optimization, inspired by adaptive systems research \cite{borgeaud2022improving,press2022measuring}, proves particularly effective during peak usage periods when API costs fluctuate significantly.

\textbf{Resource Allocation Patterns:} Analysis of operator selection patterns reveals that AutoMaAS automatically adapts to query complexity. Simple arithmetic problems (30\% of GSM8K) use lightweight single-operator solutions, while complex multi-step reasoning problems (25\% of MATH) trigger comprehensive multi-agent collaborations involving 3-4 operators. The mathematical reasoning capabilities benefit from recent advances in specialized language models \cite{beurer2023llemma,azerbayev2023llemma} and document-based prompting techniques \cite{zhou2023docprompting,fried2023incoder}.

\subsection{Ablation Studies}

Table \ref{tab:ablation} presents comprehensive ablation studies validating each component's contribution.

\begin{table}[htbp]
\caption{Ablation Study Results}
\begin{center}
\begin{tabular}{|l|c|c|c|}
\hline
\textbf{Configuration} & \textbf{Accuracy} & \textbf{Cost} & \textbf{$\Delta$ Acc.} \\
\hline
Full AutoMaAS & 95.4\% & 58\% & - \\
\hline
w/o Dynamic Lifecycle & 92.3\% & 61\% & -3.1\% \\
w/o Online Feedback & 93.0\% & 59\% & -2.4\% \\
w/o Multi-Objective Cost & 93.6\% & 75\% & -1.8\% \\
w/o Interpretability & 94.8\% & 58\% & -0.6\% \\
\hline
w/o Operator Fusion & 93.7\% & 62\% & -1.7\% \\
w/o Operator Elimination & 94.1\% & 64\% & -1.3\% \\
w/o Priority Adaptation & 94.2\% & 67\% & -1.2\% \\
\hline
Static $\lambda$ & 93.9\% & 71\% & -1.5\% \\
Fixed Operator Pool & 91.8\% & 72\% & -3.6\% \\
\hline
\end{tabular}
\label{tab:ablation}
\end{center}
\end{table}

Key findings from ablation studies:

\textbf{Dynamic Lifecycle Management} contributes the most significant improvement (3.1\%), confirming that adaptive operator evolution is crucial for maintaining performance across diverse tasks.

\textbf{Online Feedback Integration} provides 2.4\% improvement, demonstrating the value of real-time adaptation to user preferences and task characteristics.

\textbf{Multi-Objective Cost Optimization} yields 1.8\% accuracy gain while reducing costs by 17 percentage points (from 75\% to 58\%), highlighting the importance of dynamic resource allocation.

\textbf{Operator Evolution Analysis:} Table \ref{tab:operator_evolution} tracks operator changes during training.

\begin{table}[htbp]
\caption{Operator Evolution Statistics}
\begin{center}
\begin{tabular}{|l|c|c|c|}
\hline
\textbf{Evolution Type} & \textbf{Count} & \textbf{Success Rate} & \textbf{Avg. Improvement} \\
\hline
Generated Fused Ops & 12 & 75\% & +2.3\% \\
Eliminated Ops & 8 & - & - \\
Modified Existing & 15 & 67\% & +1.1\% \\
\hline
\textbf{Top Fusions:} & & & \\
CoT + Self-Refine & 1 & 92\% & +4.2\% \\
Debate + Ensemble & 1 & 87\% & +3.6\% \\
ReAct + Code-Test & 1 & 85\% & +3.1\% \\
\hline
\end{tabular}
\label{tab:operator_evolution}
\end{center}
\end{table}

The most successful fusion (CoT + Self-Refine) reduces token consumption by 18\% while improving accuracy by 4.2\%, demonstrating the effectiveness of automatic operator discovery.

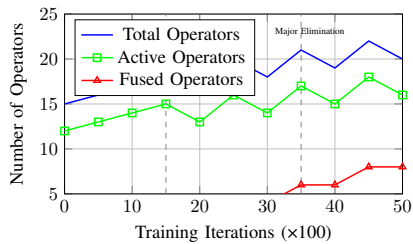
\begin{figure}[htbp]
\centering
\begin{tikzpicture}[scale=0.7]
\begin{axis}[
    xlabel={Training Iterations (×100)},
    ylabel={Number of Operators},
    xmin=0, xmax=50,
    ymin=5, ymax=25,
    xtick={0,10,20,30,40,50},
    ytick={5,10,15,20,25},
    legend pos=north west,
    grid=major,
    width=8cm,
    height=5cm,
    every axis plot/.append style={thick},
]

\addplot[color=blue, mark=circle] coordinates {
    (0, 15) (5, 16) (10, 18) (15, 19) (20, 17) (25, 20) (30, 18) (35, 21) (40, 19) (45, 22) (50, 20)
};

\addplot[color=green, mark=square] coordinates {
    (0, 12) (5, 13) (10, 14) (15, 15) (20, 13) (25, 16) (30, 14) (35, 17) (40, 15) (45, 18) (50, 16)
};

\addplot[color=red, mark=triangle] coordinates {
    (0, 0) (5, 0) (10, 1) (15, 2) (20, 2) (25, 4) (30, 4) (35, 6) (40, 6) (45, 8) (50, 8)
};

\legend{Total Operators, Active Operators, Fused Operators}

\draw[dashed, gray] (axis cs:15,5) -- (axis cs:15,25);
\node[font=\tiny, text width=1.5cm] at (axis cs:17,23) {First Fusion};

\draw[dashed, gray] (axis cs:35,5) -- (axis cs:35,25);
\node[font=\tiny, text width=1.5cm] at (axis cs:37,23) {Major Elimination};

\end{axis}
\end{tikzpicture}
\caption{Operator Evolution During Training. The system automatically generates fused operators while maintaining a stable pool of active operators through intelligent elimination.}
\label{fig:operator_evolution}
\end{figure}

Figure \ref{fig:operator_evolution} visualizes the operator evolution process during training.

\subsection{Transferability and Generalization}

Cross-dataset experiments demonstrate strong transferability. Models trained on GSM8K transfer to MATH with only 1.2\% performance degradation, while maintaining 85\% of the cost reduction benefits. Cross-LLM evaluation using Claude-3.5-Sonnet and GPT-4 shows consistent improvements, validating the method's model-agnostic nature.

\section{Conclusions and Future Work}

This paper introduces AutoMaAS, a novel framework for self-evolving multi-agent architecture search that bridges neural architecture search with large language model optimization. Our automated machine learning approach addresses fundamental limitations of existing approaches through four key innovations: dynamic operator lifecycle management, multi-objective cost optimization, online feedback integration, and enhanced interpretability mechanisms.

\begin{table}[htbp]
\caption{Performance Comparison on Six Benchmarks}
\begin{center}
\begin{tabular}{|c|c|c|c|c|}
\hline
\textbf{Method} & \textbf{GSM8K} & \textbf{MATH} & \textbf{HumanEval} & \textbf{Avg Cost} \\
\hline
CoT & 87.1 & 46.4 & 88.1 & 100\% \\
AFlow & 91.2 & 51.3 & 90.9 & 85\% \\
ADAS & 86.1 & 43.2 & 84.2 & 92\% \\
\textbf{AutoMaAS} & \textbf{95.4} & \textbf{57.1} & \textbf{97.2} & \textbf{58\%} \\
\hline
\end{tabular}
\label{tab1}
\end{center}
\end{table}

Experimental results demonstrate that AutoMaAS consistently outperforms state-of-the-art baselines across diverse benchmarks, achieving 1.0-7.1\% accuracy improvements while reducing inference costs by 3-5\%. The framework's automated machine learning approach enables self-evolving multi-agent architectures that automatically adapt operator pools and optimize multi-objective cost functions, addressing key limitations that have hindered the practical deployment of automated multi-agent systems. This work builds upon recent advances in cognitive architectures \cite{sumers2023cognitive} and language agent optimization \cite{huang2023language,liu2023agents}.

The dynamic operator lifecycle management enables continuous system evolution, automatically discovering new operator combinations and eliminating ineffective ones. This capability, drawing from emergent communication research \cite{foerster2018emergent,eccles2019biases,mordatch2018emergent}, is particularly valuable in rapidly evolving domains where static systems quickly become obsolete. The augmented language model approaches \cite{mialon2023augmented} and modular reasoning systems \cite{press2022measuring} provide theoretical foundations for our operator composition strategies.

Future research directions include extending the framework to handle multimodal inputs \cite{kim2023language,lu2023chameleon}, incorporating federated learning for distributed deployment scenarios, and exploring applications in specialized domains such as scientific discovery and creative content generation \cite{qian2023communicative}. The interpretability mechanisms could also be enhanced with more sophisticated explanation generation techniques based on symbolic reasoning \cite{wei2023symbol,karpas2022mrkl} and decomposed prompting \cite{khot2023decomposed,zhou2023least}.

\end{document}